%% file: arxiv.tex
\documentclass[10pt,twocolumn,letterpaper]{article}

\usepackage{wacv}
\usepackage{times}
\usepackage{epsfig}
\usepackage{graphicx}
\usepackage{amsmath}
\usepackage{amssymb}
\usepackage{multirow}
\usepackage{booktabs}
\usepackage{mathtools}
\usepackage{bm}
\usepackage{lipsum}
\usepackage{bbm}
\usepackage[shortlabels]{enumitem}
\usepackage[accsupp]{axessibility}  
\usepackage{pdfpages}
\usepackage{multido}

\def\wacvPaperID{331} 

\wacvalgorithmstrack   

\wacvfinalcopy 

\ifwacvfinal
\usepackage[breaklinks=true,bookmarks=false]{hyperref}
\else
\usepackage[pagebackref=true,breaklinks=true,colorlinks,bookmarks=false]{hyperref}
\fi
\usepackage[capitalize]{cleveref}
\crefname{section}{Sec.}{Secs.}
\Crefname{section}{Section}{Sections}
\Crefname{table}{Table}{Tables}
\crefname{table}{Tab.}{Tabs.}

\newcommand{\mtos}{ModelNet$\rightarrow$ShapeNet}
\newcommand{\mtosc}{ModelNet$\rightarrow$ScanNet}
\newcommand{\stom}{ShapeNet$\rightarrow$ModelNet}
\newcommand{\stosc}{ShapeNet$\rightarrow$ScanNet}

\newcommand{\pl}{pseudo-labels}

\newcommand{\st}{self-training}
\newcommand{\sd}{self-distillation}
\newcommand{\graph}{$\mathcal{G}$}
\newcommand{\std}{$\Phi$}
\newcommand{\thc}{$\tilde{\Phi}$}
\newcommand{\pcd}{$x$}
\newcommand{\pcdas}{$x''$}
\newcommand{\pcdaw}{$x'$}
\newcommand{\desc}{$g$}
\newcommand{\desca}{$\tilde{g}$}

\begin{document}

\title{Self-Distillation for Unsupervised 3D Domain Adaptation}

\author{Adriano Cardace\qquad Riccardo Spezialetti\qquad Pierluigi Zama Ramirez\qquad \\
Samuele Salti\qquad Luigi Di Stefano\\
Department of Computer Science and Engineering (DISI)\\
University of Bologna, Italy\\
{\tt\small \{adriano.cardace2, riccardo.spezialetti, pierluigi.zama\}@unibo.it}
}

\maketitle
\thispagestyle{empty}

\begin{abstract}
Point cloud classification is a popular task in 3D vision. However, previous works, usually assume that point clouds at test time are obtained with the same procedure or sensor as those at training time. Unsupervised Domain Adaptation (UDA) instead, breaks this assumption and tries to solve the task on an unlabeled target domain, leveraging only on a supervised source domain.
For point cloud classification, recent UDA methods try to align features across domains via auxiliary tasks such as point cloud reconstruction, which however do not optimize the discriminative power in the target domain in feature space.
In contrast, in this work, we focus on obtaining a discriminative feature space for the target domain enforcing consistency between a point cloud and its augmented version.
We then propose a novel iterative self-training methodology that exploits Graph Neural Networks in the UDA context to refine pseudo-labels.
We perform extensive experiments and set the new state-of-the art in standard UDA benchmarks for point cloud classification. Finally, we show how our approach can be extended to more complex tasks such as part segmentation.
\end{abstract}

\input{sections/00-introduction}
\input{sections/01-related}
\input{sections/02-method}
\input{sections/03-experiments}
\input{sections/04-conclusions}

{\small
\bibliographystyle{ieee_fullname}
\bibliography{egbib}
}

\newpage\phantom{Supplementary}
\multido{\i=1+1}{4}{
\includepdf[page={\i}]{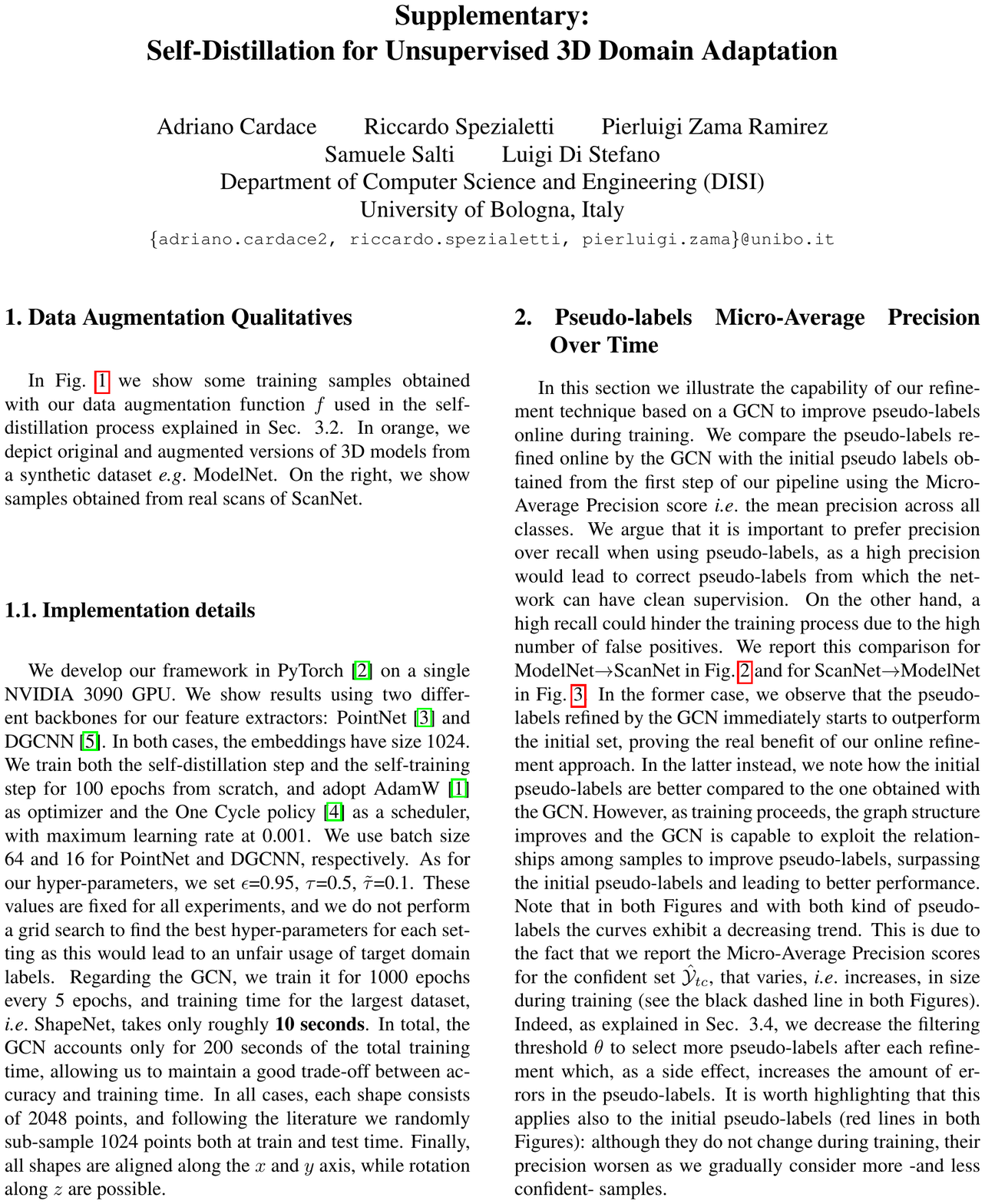}
}

\end{document}

%% file: sections/00-introduction.tex
\section{Introduction}
\label{sec:intro}
In recent years, point cloud classification has received a lot of attention due to its relevance to many practical applications such as scene understanding, augmented/mixed reality, robotics and autonomous driving \cite{Geiger2012CVPR,sun2020scalability}. Deep learning brings the promise of data-driven solutions to this problem and a variety of deep architectures have arisen in response to this challenge \cite{Qin2019pointdan,qi2017pointnet++,liu2020closer,yan2020pointasnl,xu2020grid,liu2019relation,hua2018pointwise,wang2019dynamic,thomas2019kpconv}.

The success of these approaches goes hand in hand with the availability of large datasets containing labelled shapes \cite{modelnet,chang2015shapenet}. However, most existing annotated datasets concern clean and occlusions-free CAD shapes, but deep models trained on these objects drastically fail when facing data with different characteristics.
This is the case, in particular, of models trained on synthetic CAD data and then tested on point clouds obtained with real sensors, where parts of the object may be missing due to occlusions and measurements are corrupted by noise.
Here comes to help Unsupervised Domain Adaptation (UDA), which pursues solving a supervised learning task in a \textit{Target} domain, $\mathcal{T}$, where data come without labels,  by  leveraging on labeled data available in a \textit{Source} domain, $\mathcal{S}$.
In the last couple of years, an increasing number of papers \cite{cardace2021refrec,Qin2019pointdan,achituve2021self,alliegro2021joint,Zou_2021_ICCV} have addressed UDA for point cloud classification, with popular synthetic datasets of CAD models, such as ModelNet40 \cite{wu20153d} or ShapeNet \cite{chang2015shapenet}, and real datasets such as ScanNet \cite{dai2017scannet}.
\begin{figure}[t]
    \centering
    \includegraphics[scale=0.4, trim={6cm 4.5cm 8.5cm 4.5cm}, clip]{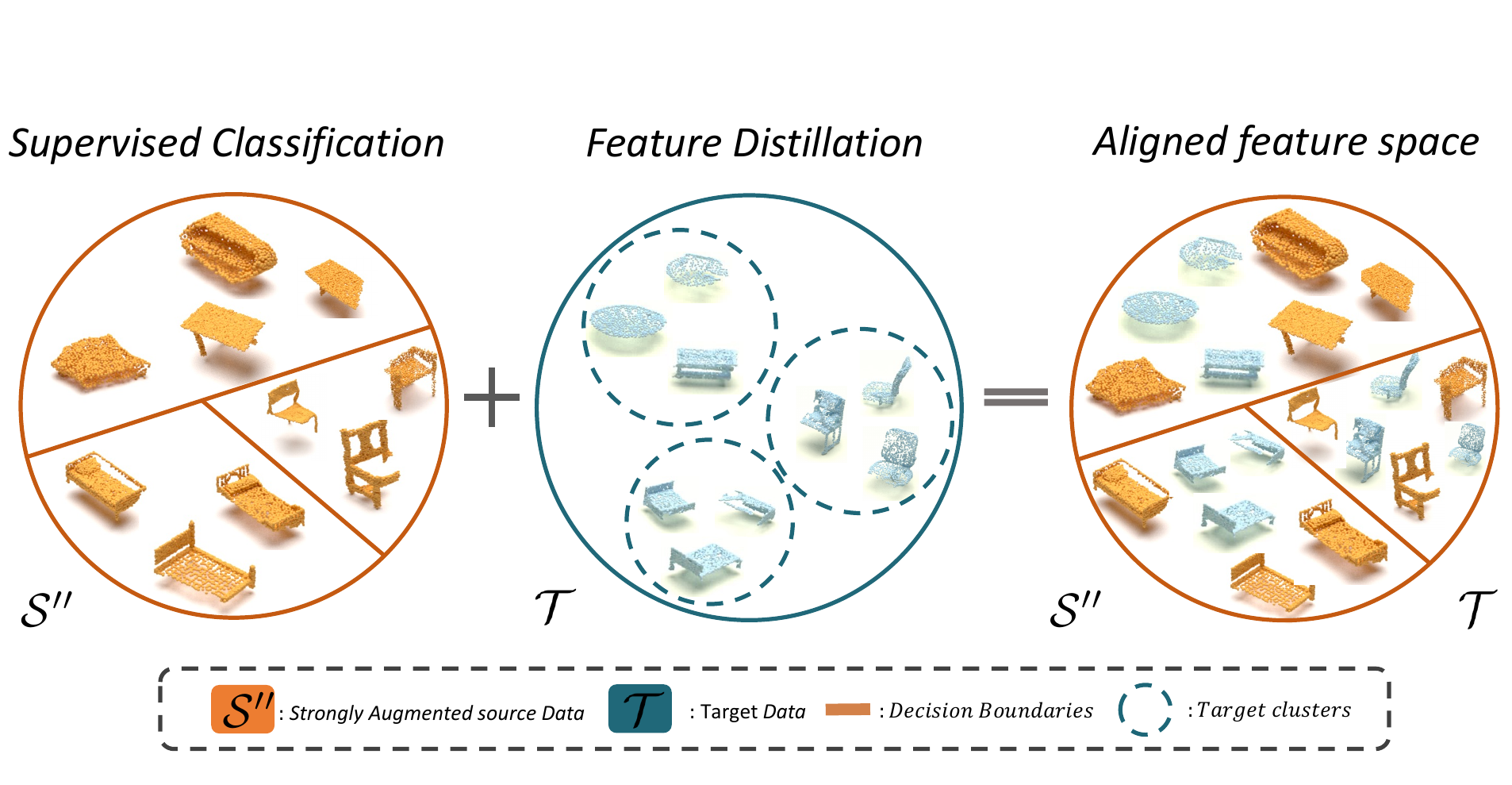}
    \caption{
    Proposed UDA method. We combine a supervised training of strongly augmented source data with a \sd{} approach that aims at clustering target shapes unsupervisedly. The combination of these two approaches leads to an alignment in feature space across domains.
    }
    \label{fig:teaser}
\end{figure}
The main line of research focuses on learning an effective feature space for the target domain by means of auxiliary tasks such as point cloud reconstruction \cite{achituve2021self,Shen_2022_CVPR}, 3D puzzle sorting \cite{alliegro2021joint} and rotation prediction \cite{Zou_2021_ICCV}.
These tasks are refereed as auxiliary since they do not directly solve the main task, but at the same time, they are useful to learn features for the target domain without the need of annotations. %
Although such techniques considerably improve over the baseline (i.e., training only on source data), the design of such tasks is not trivial and typically lead to sub-optimal solutions.
It requires identifying one that can drive the network to learn representations \textit{discriminative} enough to perform classification in the target domain effectively.
Despite the fact they force some degree of alignment between the features computed on objects from the two domains, such auxiliary tasks do not explicitly steer the network to learn discriminative representations amenable to classification in the target domain.
For instance, if we train a network to reconstruct shapes, we will get similar point cloud embeddings for similar 3D shapes. However, two point clouds could represent objects that, though similar in shape, do belong to different categories, \eg{}, a cabinet and a bookshelf.
As a consequence, relying on reconstruction to perform domain adaptation can align features between the two domains, with similar shapes embedded close one to each other regardless of their domain, but the decision boundaries learnable from the labeled source samples may not discriminate effectively between target samples belonging to different classes. This is also shown in \cite{achituve2021self}, where a simple denoising auto encoder for point clouds only slightly improves performance over the baseline.
We reckon that similar considerations apply to the other auxiliary tasks proposed in the literature as they pursue cross-domain feature alignment based on a learning objective that does not ensure cross-domain class discriminability. We support this claim by comparing our proposal with previous works in the experimental section.

In this paper instead, we take inspiration from a recent self-supervised approach, DINO \cite{caron2021emerging}, to learn more  discriminative representations for the target domain by constraining a sample and a strongly augmented version of itself to be classified similarly. This is typically achieved trough \sd{}, a methodology where the output of a neural network is compared with the output obtained from a mean teacher \ie{} a temporal exponential moving average of the weights of the network itself (EMA) \cite{ema}.
As shown in \cite{caron2021emerging} this training methodology allows for clustering together samples of the same class.
However, differently from DINO, we apply \sd{} for the first time in the 3D UDA context, where samples are point clouds, and the main goal is to reduce the gap between representations of two different domains rather than only learning a well-clustered feature space for a single domain.
We believe that \sd{} is particularly suited for point cloud domain adaptation due to peculiar 3D data augmentations such as translation, occlusion and point-wise noise that can easily bridge the gap between source and target domain.
By exploiting such augmentations to strongly augment source data and by enforcing inter-class discriminability for the target domain via \sd{}, we are able to obtain a shared aligned feature space across domains. The overall idea is illustrated in \cref{fig:teaser}).
Moreover, one major limitation of DINO that hinders its wide adoption is mode collapse \cite{caron2021emerging}, and previous works usually adopt multiple tricks and hyper-parameters such as clustering constraints \cite{caron2018deep}, predictor \cite{grill2020bootstrap} and contrastive losses \cite{wu2018unsupervised} that are difficult to apply and tune in other contexts.
In this work, we show how this paradigm can be applied without such tricks to UDA for point cloud classification, where mode collapse is prevented by simultaneously training a classifier on labelled source data that inherently separates the features space according to semantic categories.

In the second step of our proposal, following recently published works in the field \cite{cardace2021refrec,Zou_2021_ICCV,Shen_2022_CVPR,Fan_2022_CVPR}, we make use of \textit{\st{}}, an iterative methodology that exploits the predictions of a pre-trained model (\textit{\pl{}}) to provide partial supervision on the target domain as well.
However, \pl{} are noisy and their naive use typically leads to overfitting of the dominant classes of the source domain as shown in \cite{Zheng2020,Zou2019}. The strategy proposed by \cite{cardace2021refrec} to refine them requires offline training of an additional network for this purpose and the definition of hand-crafted rules based on k-NN queries, limiting its general applicability, while \cite{Zou_2021_ICCV} adopted a standard procedure borrowed from the 2D world \cite{zou2018unsupervised}.
As a further contribution of our work, we take a different path, and propose to use Graph Neural Networks (GNNs) \cite{survey2021} to refine \pl{} online during self-training.
Our main intuition is that by using a GNN, \pl{} are obtained by considering relationships between all target samples in the dataset rather than on single samples in isolation. This allows for reasoning at the dataset-level and enables to correct misclassified samples and thus refine \pl{}.
Moreover, the target feature space is clustered thanks to the \sd{}, thus each node of the graph is likely to be connected to samples of the same category. Hence, the GNN can improve the \pl{} by reasoning on a neighborhood of samples sharing the same class. This procedure can be done online during training with the graph structure evolving over time, thus avoiding \pl{} overfitting.

Project Page at \url{https://cvlab-unibo.github.io/FeatureDistillation/}. In short, our contributions can be summarised as follows:
\begin{enumerate}
    \item we propose the first approach in UDA for point clouds classification that exploits self-distillation to learn effective representations to classify point clouds in the target domain;
    
    \item we show a novel strategy to use GNNs in UDA for point clouds classification. It enables online refinement of \pl{}, which reduces risks of overfitting, and it is conducive to effective self-training;
    
    \item we extensively test our framework on standard benchmarks \cite{Qin2019pointdan}
    and establish new state-of-the art results. Furthermore, we show how our approach can be generalized to the challenging task of part segmentation.

    \end{enumerate}

%% file: sections/01-related.tex
\section{Related Work}
\label{sec:related}

%

\textbf{Unsupervised 3D Domain Adaptation. }
Unsupervised Domain Adaptation emerged in the last years as a technique able to alleviate the domain shift when training a neural network on a source domain (e.g., synthetic simulations) and test in an unlabeled different, but related target domain (e.g., real data).
UDA has a rich literature in the 2D world, and a noticeable amount of work as been conducted for image classification \cite{ganin2015,Bousmalis2016,Long2017,NIPS2016_ac627ab1,Sun2016DeepCC,Tzeng2017}, semantic segmentation \cite{bdl,adaptsegnet,maxsquare,pmlr-v80-hoffman18a} and object detection \cite{Chen2018,Wang_2019_CVPR,Wang_2019_CVPR}.
PointDAN \cite{Qin2019pointdan} has been the first work to address point cloud classification in the UDA context; they leverage on the well known Maximum Classifier Discrepancy (MCD) \cite{Saito2018MaximumCD} to achieve alignment in feature space.
Differently, \cite{alliegro2021joint,achituve2021self,Zou_2021_ICCV} exploit Self-Supervised Learning (SSL) to run additional tasks on both domains.
\cite{cardace2021refrec} also leverages on point cloud reconstruction, but uses it to refine \pl{}. Despite its effectiveness, their pipeline is rather complex and based on ad-hoc k-NN queries.
The main difference with these works is in the way we exploit 3D transformations: while they use transformations such as rotations or point-wise jittering to solve an additional task in a self-supervised fashion, we use these augmentations in input space to design a novel distillation approach that pushes the network to learn a discriminative feature space for the target domain.

\textbf{Self-training.}
Self-training \cite{zou2018unsupervised} is a common technique used in Domain Adaptation to assign noisy annotations to target samples \ie{} \pl{} \cite{lee2013pseudo}, so that partial supervision can be provided to learn the distributions of the target domain.
Pseudo-labels are often fairly inaccurate and many methods have been proposed to address this issue for UDA for image classification \cite{Gu_2020_CVPR,ChenWKM20,ShinWPK20}, semantic segmentation \cite{mei2020instance,bdl,ltir}, and object detection \cite{Kim_2019_ICCV,Wang2021} by either filtering or refining pseudo labels.
The potential of \st{} has also been showed for point clouds classification in \cite{cardace2021refrec} where \pl{} are refined by an auxiliary reconstruction task.
We also leverage on this powerful technique and propose for the first time to refine \pl{} using Graph Neural Networks in the UDA context.

\textbf{Knowledge distillation.}
%
The case in which soft \pl{} rather than hard labels are used is typically denoted as \textit{Knowledge distillation} \cite{hinton2015distilling}.
Although distillation has been originally introduced to boost performance of small neural networks, recent works revisited knowledge distillation as way to learn robust features for better initialization or image retrieval \cite{caron2021emerging}. In particular, DINO \cite{caron2021emerging} proposed a novel framework able to learn robust features exploiting augmented versions of the same images of a given domain.
Inspired by DINO\cite{caron2021emerging}, we propose to apply such paradigm to tackle the UDA scenario for 3D objects. Indeed we aim at showing that self-distillation can be applied exploiting 3D augmentations, and more importantly, that we can design such learning protocol to reduce the gap between a source and a target domain.

\textbf{Graph Neural Networks (GNNs). }
Recent GNNs models \cite{kipf2016semi,velickovic2018graph,defferrard2016convolutional} have emerged as powerful architectures for graph-structured data, covering a large spectrum of applications: social analysis \cite{li-goldwasser-2019-encoding,DeepInf}, drug discovery \cite{NIPS2015_f9be311e} and recommendation systems \cite{2019Recommendation,2019Session-Based}.
The rich literature on 2D semi-supervised learning \cite{kipf2016semi,velickovic2018graph,nrgnn,2019Li,UniMP,DeepInf} already provides many works that leverage on GNNs to assign labels on unlabelled nodes.
However, all these works, assume a small amount of \textit{perfectly} labelled nodes in the graph for each class, while this assumption does not hold in the UDA scenario.
To the best of our knowledge, \cite{Ma_2019_CVPR} is the only paper that addresses UDA for image classification using GNNs. They focus on extracting complementary features through a GNN to be combined with features obtained with a classical Convolutional Neural Network (CNN).
Other works such as \cite{Graph_Matching} and \cite{Ding2018GraphAK} instead, exploit graph structures (not GNNs) with hand-crafted label propagation algorithms to achieve adaptation.
Differently, we propose the usage of GNNs to obtain new \pl{} while self-training, to avoid overfitting and to allow an iterative refinement of them to converge to better adaptation performances.
Moreover, we are the first to show their effectiveness in the case of UDA for 3D shape classification.

%% file: sections/02-method.tex
\section{Method}
\begin{figure*}[t]
	\centering
	\includegraphics[scale=0.35, trim={0.5cm 3.5cm 0.5cm 3.5cm}, clip,]{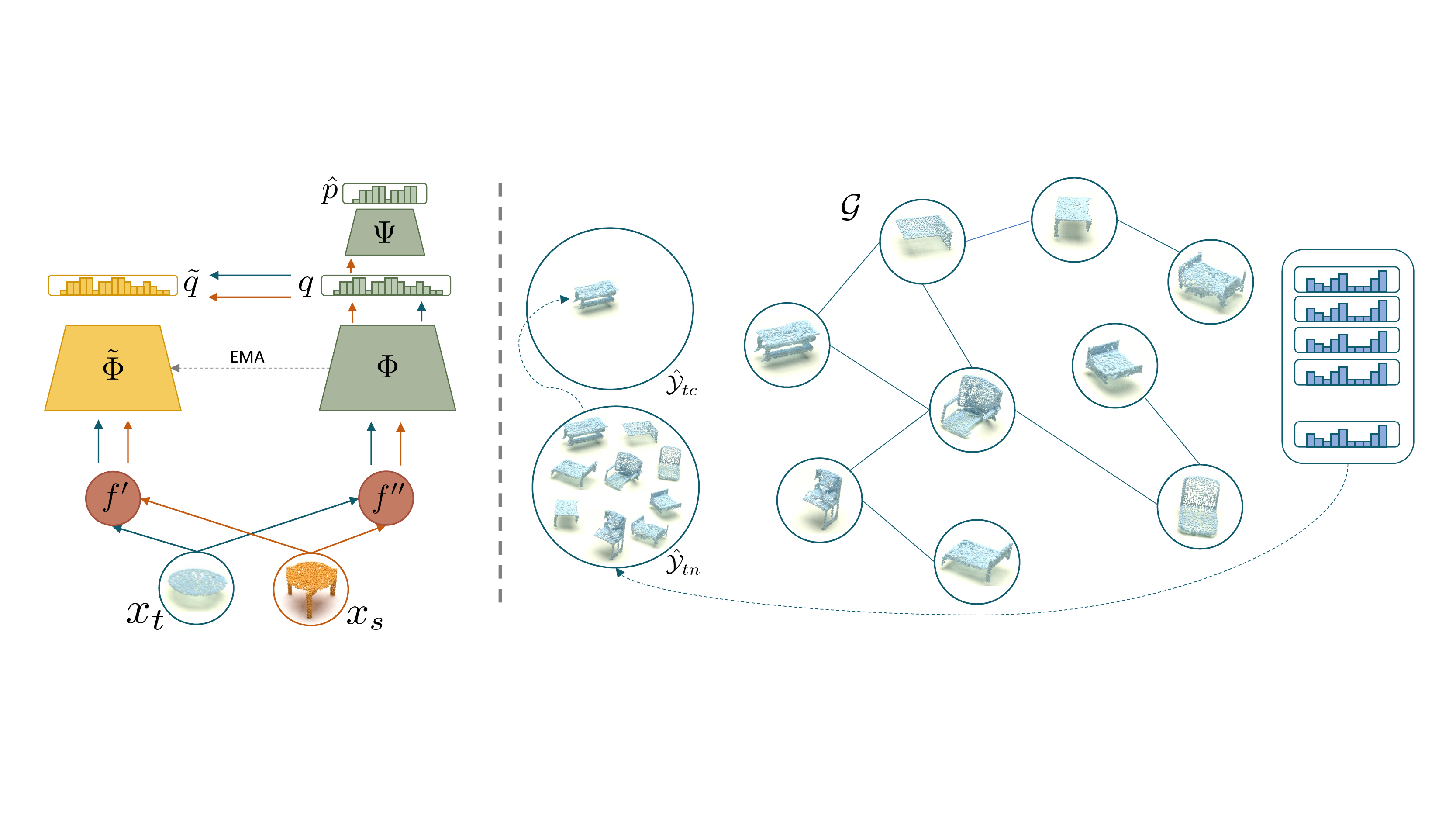}
	\caption{Illustration of our framework. \textbf{Left}: weakly and strongly augmented point clouds are generated with two transformation functions $f'$ and $f''$ for both domains. The weakly augmented shapes are fed to an exponential moving average (EMA) encoder, the teacher $\tilde{\Phi}$, while the strongly augmented are processed by the student $\Phi$. A consistency loss is applied between the corresponding embeddings. \textbf{Right}: the whole target dataset is processed by a GCN $\mathcal{G}$ online during self-training to iteratively refine and update \pl{}}
	\label{fig:framework}
\end{figure*}

\label{sec:method}
Our framework is divided into two main steps: \sd{} (\cref{subsec:fd}) and \textit{\st{} with \pl{} refinement} (\cref{subsec:plext} and \cref{subsec:st}). The overall pipeline is depicted in \cref{fig:framework}. We start introducing the notation and a brief review of the basic concepts about GNNs.

\subsection{Preliminaries}
\label{subsec:notation}
\textbf{Notation.}
In this paper, we consider UDA for point cloud classification, \ie{} given a point cloud with $N$ elements $x \in \mathbb{R}^{N\times3}$ we aim at learning a neural network $\Omega: x \rightarrow [0, 1]^K$ that takes an input example $x$ and produces a $K$-dimensional vector representing the confidence scores for  $K$ classes. Such a point cloud classifier consists of two components: $\Omega=\Phi \circ \Psi$. The first is a feature extractor network, $\Phi: \mathbb{R}^3 \rightarrow \mathbb{R}^D$, producing $g \in \mathbb{R}^D$, \ie{} a $D$-dimensional global feature descriptor for the shape, the second is small MLP $\Psi: \mathbb{R}^D \rightarrow \mathbb{R}^K$ followed by a softmax operator which maps \desc{} to a vector of confidence scores $\hat{p} \in [0,1]^K$. Finally, the class predictions can be obtain by the argmax operator $\Lambda: \mathbb{R}^K \rightarrow \mathcal{Y}$. As it is peculiar in UDA settings, we have at our disposal a source domain with labels $\mathcal{S} = {\{(x_s^i \in \mathcal{X}_s, y_s^i \in \mathcal{Y}_s)\}_{i=1}^{n_s}}$, and a target domain $\mathcal{T} = \{x_t^j \in \mathcal{X}_t\}_{j=1}^{n_t}$, where the point clouds are unlabeled. Our objective is to obtain a classifier able to make correct predictions on $\mathcal{T}$.

\textbf{Background on GNNs.}
Graph Neural Networks (GNNs) are models designed to process graphs, \ie{} sets of nodes that are optionally joined
one  to another  by edges representing relationships. GNNs are a powerful tool to process unstructured data thanks to their ability of updating the representations of each node by aggregation of information from the neighbouring nodes. An undirected graph $\mathcal{G}$ is represented as a tuple $(\mathcal{V}, \mathcal{E})$, where $\mathcal{V}$ is the set of $N$ vertices $v_{i} \in \mathcal{V}$, and $\mathcal{E}$ is the set of edges.
The graph topology is determined by the adjacency matrix $A \in \mathbb{R}^{N\times N}$, with $A_{i,j}=1$ if two nodes $i$ and $j$ are connected.
Among the many architectures of GNNs \cite{survey2021}, in this work we adopt the Graph Convolutional Networks (GCNs) \cite{kipf2016semi}, which propagate the information between each layer according to the following propagation rule:
\begin{equation}
  \textstyle
  H^{(l+1)}= \sigma\!\left(\tilde{D}^{-\frac{1}{2}} \tilde{A}\tilde{D}^{-\frac{1}{2}}H^{(l)} W^{(l)} \right) \,
\label{eq:gcn-layer}
\end{equation}
where $\tilde{A} = A + I$ represents the adjacency matrix with self-connections, $I$ is the identity matrix, $\tilde{D}_{ii} = \sum_j \tilde{A}_{ij}$ acts as a scaling factor and $W^{(l)}$ is a layer-specific trainable weight matrix. 
The aggregation rule is followed by a non-linear activation function $\sigma(\cdot)$ such as ReLU.
Note that matrix $H^{(l)}$ deals with the $l$-th layer of the network, each row $i$ representing the feature vector of a node $v_i \in \mathcal{V}$ in that layer. We refer the reader to \cite{survey2021} for a more detailed discussion. 

\subsection{Self-distillation}
\label{subsec:fd}
In this section, we present the self-distillation module that we use in both steps of our pipeline. 
The purpose of this component is to distill good features for the target domain unsupervisedly, so that a discriminative feature space directly useful for classification can be learned even though no direct supervision is available in $\mathcal{T}$.
Our main intuition, is that learning a clustered feature space that minimises the distance among variations of the same cloud from the target domain, while simultaneously learning decision boundaries amenable to classification thanks to a carefully augmented source domain, is key to obtain good \pl{} to be deployed in the self-training process. 
Indeed, without enforcing compactness in the feature space, it is more likely that, due to the domain gap, target samples are spread across the different categories defined by the decision boundaries of the classifier. This is undesirable since it would lead to excessive noise in \pl{}.

To achieve our goal, we use two data augmentation functions $f', f'': \mathbb{R}^{N\times3} \rightarrow \mathbb{R}^{N\times3}$ that take as input a point cloud \pcd{} and return a weakly augmented (\pcdaw{}) and a strongly augmented point cloud (\pcdas{}) respectively.
Then, we adopt a \sd{} paradigm, where we train a student encoder \std{} to match the output of a teacher encoder \thc{}. 
In particular, we match two global shape descriptors, $\tilde{g} = \tilde{\Phi} ($\pcdaw{}$)$ and $g = \Phi($\pcdas{}$)$, computed by feeding a weakly augmented point cloud \pcdaw{} to 
the teacher and the strongly augmented version \pcdas{} to the student.

By taking inspiration from \cite{caron2021emerging}, we design the student and the teacher to output probability distributions over $D$ dimensions, denoted by $q$ and $\tilde{q}$, respectively. These probabilities can be obtained by normalizing the output of the two encoders, \ie{} \desc{} and \desca{}, with a softmax function: 
\begin{equation}
\label{eq:softmax}
\begin{aligned}
  q(g, \tau) &= \frac{\exp(g / \tau)}{\sum_{d=1}^D \exp(g^{(d)} / \tau)}, \\
  \tilde{q}(\tilde{g}, \tilde{\tau}) &= \frac{\exp(\tilde{g} / \tilde{\tau})}{\sum_{d=1}^D \exp(\tilde{g}^{(d)} / \tilde{\tau})}
\end{aligned}
\end{equation}                                                         
where $\tau>0$ and $\tilde{\tau}>0$ are the two temperature parameters
which control the sharpness of the output distributions for the student and the teacher, respectively. Differently from \cite{caron2021emerging}, we don't require any complex scheduling for the temperature parameters, and we just empirically set them to 0.5 by observing the model performance on the source domain for the \mtosc{} experiment and set it to the same value for all the others.
To force the embedding of the augmented point cloud to match that computed for the original one, we minimize the cross-entropy:
\begin{equation}
\label{eq:dino_loss}
\mathcal{L}_{sd}(\tilde{g}, g) = - \tilde{q}(\tilde{g}, \tilde{\tau}) \log q(g, \tau)
\end{equation}
by running backprop on the student network $\Phi$, while the weights of the teacher are updated by computing an exponential moving average of those of the student. Please note that both networks share the same architecture but have different weights. We employ an EMA as a teacher network since it is a convenient way to provide robust and stable features throughout the training process without the need of training another network \cite{ema,he2020momentum}. 

\textbf{Data augmentation and transformation functions.}
To implement $f'$ and $f''$, we use a set of common data augmentation techniques for point clouds such as: jittering, elastic deformation \cite{Zhou2018}, scaling along the three axis.
More specifically, to obtain the weakly augmented point cloud \pcdaw{}, we only use jittering, while for the the strongly augmented \pcdas{}, we employ all the above transformations. Additionally, when performing synthetic-to-real adaptation, we also include random point removal \cite{spezialetti2020learning}. We refer to the supplementary material for some qualitative examples.

Interestingly, the same 3D transformations can be used to simulate the target distribution given source data. In fact, although it is not possible to exactly predict the shift between two domains, one can approximate the nuisances that affect the target data through aggressive data augmentation. 
For example, when performing UDA between different synthetic domains, shapes may have similar geometric elements but with different style \cite{lun2015elements,xu2010style}, which can be mimicked by object distortions or elongation and scaling.
Similarly, when moving from a synthetic domain to a real one, it is reasonable to assume that shapes within the same class will appear similar to CAD models but will have missing components due to occlusions, and point coordinates will be affected by the noise originated in the acquisition process. 
Therefore, as shown in \cref{fig:framework} (left), at training time we augment the source data re-utilizing the transformation function $f''$, with the goal of minimising the gap between the two domains in the input space and seamlessly obtain a better alignment also in the feature space. Applying such well designed augmentations to source data combined with our distillation technique is beneficial to the student model. 
Intuitively, by distillation we aim at clustering target samples, while by data augmentation we force source clusters, naturally obtained with a classification loss, to be aligned with the target ones.

\subsection{Pseudo-labels initialization}
\label{subsec:plext}
In the first step of our method, we exploit the self-distillation module presented in the previous section to obtain an initial set of \pl{} for the target domain.
In particular, as shown in \cref{fig:framework} (left), we train a classifier  $\Omega=\Phi \circ \Psi$ on top of the student feature extractor and fed with augmented source data. We use the cross entropy loss:
\begin{equation}
\label{eq:ce}
\mathcal{L}_{ce}(x''_s, y_s) = - y_s \log \Omega(x''_s)\\
\end{equation}
Simultaneously, we feed to the encoder $\Phi$ batches of source and target point clouds strongly augmented with the transformation function $f''$, while $\tilde{\Phi}$ receives the weakly augmented versions, and minimise \cref{eq:dino_loss} to learn the desired clustered feature space for the data of the target domain.
After training, the initial set of \pl{} is computed by feeding each target sample $x^i_t$ into $\Omega$ and selecting the class with the highest confidence score: $\hat{y}^i_t = \Lambda ( \Omega(x^i_t))$.

\subsection{Self-training and \pl{} refinement}
\label{subsec:st}
In the second step, we exploit and refine the previously obtained \pl{}. We do this by alternating self-training and refinement in an iterative procedure.

\textbf{Self-training.}
In this step we train our classifier $\Omega=\Phi \circ \Psi$ leveraging \pl{}, starting from scratch if it is the first iteration.
To do so, we first split the pairs of target samples and associated pseudo-labels $(x^i_t, \hat{y}^i_t)$  into two disjoint sets, \ie{} $\hat{\mathcal{Y}}_{tc}$ and $\hat{\mathcal{Y}}_{tn}$, associated with confident and non-confident \pl{}, respectively, and  with the former initialized to the empty set.
The sets will be useful to realize the iterative procedure outlined at the end of the section.
We then train $\Phi$ and $\Psi$ using self-distillation and supervision for both domains, with supervision for the target coming from \pl{}:
\begin{equation}
    \label{eq:total_loss}
    \begin{aligned}
        \mathcal{L} &= - \mathcal{L}_{ce}(x''_s, y_s) - \lambda \mathcal{L}_{ce}(x'_t, \hat{y}_t)
        - \mathcal{L}_{sd}(x'', x') \\
    &\text{where} \quad \lambda =
        \begin{cases}
            1, & \hat{y}_t \in \hat{\mathcal{Y}}_{tc} \\
            0.2, & \hat{y}_t \in \hat{\mathcal{Y}}_{tn}
        \end{cases}.
    \end{aligned}
\end{equation}
Note that, as in the previous step, $\mathcal{L}_{sd}$ acts on both domains. 
We provide a sensitivity analysis in the supplementary material for $\lambda$ when $\hat{y}_t \in \hat{\mathcal{Y}}_{tn}$ to show that our framework is not sensitive to this hyper-parameter.

\textbf{Refinement.}  
Naively using \pl{} as done in the previous step typically leads to ignoring the classes that are underrepresented in the source domain and to obtain sub-optimal performance on the target domain due to noise in the \pl{} \cite{mei2020instance,Shin2020}. Hence, we run self-training only for a few epochs and then refine the \pl{} exploiting a GCN. Our intuition is that, by leveraging on a global view of the target dataset, the GCN can better disambiguate hard cases compared to the initial \pl{} provided by the classifier, that, on the other hand, takes its decision on each input sample in isolation. 
For instance, even if few samples of a rare class are tightly connected (i.e. node with high degrees), it is likely for their confidence  to be  high as in their neighbourhood only nodes with the same class are present.  
The role of the GCN is therefore twofold: it corrects \pl{}; it decides which \pl{} should be considered confident and thereby moved from $\hat{\mathcal{Y}}_{tn}$ into $\hat{\mathcal{Y}}_{tc}$. 
We obtain the graph $\mathcal{G}$ by considering all samples in the target domain, as shown in \cref{fig:framework} (right), and we build the adjacency matrix $A$ based on the cosine similarity between the global shape embeddings $g$:    
\begin{equation}
\label{eq:adjacency}
\begin{aligned}
    A_{i,j} &=
    \begin{cases}
        1, & \frac{\langle g_i, g_j \rangle}{\| g_i \| \| g_j \|}>\epsilon \\ 
        0, & otherwise
    \end{cases}            
\end{aligned}
\end{equation}
with $\epsilon$ being a similarity threshold empirically set to 0.95 so that the node degree (the average number of neighbours for each node of the graph) is roughly 10. We provide in the supplementary material a sensitivity study of this hyper-parameter, showing that our framework is insensitive w.r.t. to the node degree.
This is necessary for memory constraints, as the required memory to train a GCN is highly affected by this hyper-parameter.
Inspired by \cite{UniMP}, we equip each node in \graph{} with the embedding $g$ as well as with the prediction provided by the classifier $\Omega$, \ie{} the vector $\hat{p}$. These two pieces of information provide the GCN with cues concerning both the geometric structure as well as the semantic class of the object. For example, it may be the case that two point clouds have similar embeddings and yet belong to different classes. This occurs frequently when considering a real domain, where an occluded chair with a missing back could easily be misclassified as a table or the back itself with missing legs can be confused for a monitor.
Hence, providing the GCN with the additional information on the probability distribution among the $K$ classes can help it attaining more accurate \pl{} for target samples featuring similar embeddings. 
Then, we compute the input to the GCN as
\begin{equation}
    H^{(0)} = \Phi(X_{t}) + \Omega(X_{t})W_{D}
\end{equation}
where $X_{t}$ is the set of all target samples and $W_{D} \in R^{K\times D}$ is a learnable projection matrix that projects the output distribution over $K$ classes in a $D$-dimensional space.

Afterwards, following \cref{eq:gcn-layer}, we stack three graph convolutional layers where the last acts as a node classifier that returns a matrix of size $n_t\times K$. The GCN is optimized with a classical cross-entropy loss computed over all target samples, $\hat{\mathcal{Y}}_{tn} \cup \hat{\mathcal{Y}}_{tc}$, without taking into account the confidence on their pseudo-labels.
It is worth noticing that, the predictions $\Omega(X_t)$, \ie{} part of the input to the GCN, do not necessarily match the $\hat{\mathcal{Y}}_{tn} \cup \hat{\mathcal{Y}}_{tc}$ pseudo-labels.
However, the GCN can just learn to output the same probability vector $\Omega(X_{t})$, discarding part of the input features \cite{UniMP}, and consequently failing in generalizing at test time due to label leakage.
Hence, we randomly mask (\ie{} set to zero) 20\% of the inputs $\Omega(X_{t})$ at training time.

Finally, after training, we exploit the GCN to extract confident samples, \ie{} the top $\theta$ predictions for each class, update the corresponding \pl{} with the output of the GCN, and move them from $\hat{\mathcal{Y}}_{tn}$ into $\hat{\mathcal{Y}}_{tc}$.

\textbf{Iterative training.}
We argue that the topology of the graph highly influences the output of the GCN. As the encoder improves its embeddings with multiple rounds of self-training, also thanks to the self-distillation process, \pl{} become better and better since the graph structure improves. Hence, we plug the previous steps into an iterative learning process, where we repeat:
\begin{enumerate}[a)]
    \item self-train with \cref{eq:total_loss} $\Phi$ and $\Psi$ for $e$ epochs using self-distillation 
    and supervision for both domains, with supervision for the target coming from \pl{}; 
    \item build $\mathcal{G}$ and train the GCN to refine \pl{};
    \item update current \pl{}, moving the top $\theta$ predictions of the GCN for each class from $\hat{\mathcal{Y}}_{tn}$ to $\hat{\mathcal{Y}}_{tc}$.
\end{enumerate}
To gradually increase the size of $\hat{\mathcal{Y}}_{tc}$, $\theta$ starts from 0 and grows to 1 to include more and more samples during training.
At test time, the GCN as well as the teacher encoder $\tilde{\Phi}$ can be simply discarded, with $\Omega$ being the only network required to perform inference. Although the GCN can potentially be used at test time to obtain better performance, we discard it as this would introduce additional requirements such as keeping the whole training set in memory, and computing the neighborhood of each test sample.

%% file: sections/03-experiments.tex
\section{Experiments}
To show the effectiveness of out method, we compare against state-of-the-art methods for UDA for point cloud classification such as \cite{cardace2021refrec, Shen_2022_CVPR, Fan_2022_CVPR}, using two different backbones for our feature extractors: PointNet \cite{qi2017pointnet} and DGCNN \cite{wang2019dynamic}.
Furthermore, we compare with a baseline \ie{} a simple model trained only on the source domain without any adaptation, and an oracle model, which instead assumes to have all target data available. The former constitutes the lower-bound in terms of performance, while the latter is considered as the upper-bound since all target data can be utilized.
Finally, we also conducted an experiment on the challenging task of part segmentation to show how our method can be extended to different tasks than point cloud classification. In this case, we adopt the setting introduced in \cite{alliegro2021joint}, which is the only method performing adaptation on such task for the synthetic-to-real scenario.

\label{sec:experiments}

\textbf{Datasets}. The standard dataset used for UDA for point cloud classification is PointDA-10 \cite{Qin2019pointdan}, which consists of three subsets that share the same ten classes of three popular point clouds classification datasets: ShapeNet \cite{chang2015shapenet}, ModelNet40 \cite{modelnet} and ScanNet \cite{dai2017scannet}.
This allows to define six different scenarios that involve
synthetic-to-synthetic, synthetic-to-real and real-to-synthetic adaptation.
ModelNet-10 consists of 4,183 training and 856 testing point clouds, that are extracted from synthetic 3D CAD models. Similarly, ShapeNet-10 features synthetic data only. It is the largest and most varied among the three datasets, and it comprises 17,378 training and 2,492 testing samples.
Lastly, ScanNet-10 is the only real datasets, and it consists of 6,110 and 1,769 training and testing point clouds, respectively. It has been obtained from multiple real RGB-D scans. For this reason, it exhibits several forms of noise such as errors in the registration process and occlusions.
Differently from point classification, there is no established setting for part segmentation in the literature and we refer to \cite{alliegro2021joint} as a reference since it is the only work performing synthetic-to-real adaptation from ShapeNetPart \cite{shapenetPart} to ScanOBJ-BG \cite{scanobj}. The task is solved only for the \textit{chair} class, which comprises 4 components to segment: \textit{Seat, Back, Base, Arm}.

\begin{table}[t]
  \centering
  \setlength{\tabcolsep}{0.8mm}
  \scalebox{0.54}{
  \begin{tabular}{p{8em}|ccccccc}
      \multicolumn{1}{c}{} &       &       &       &       &       &       &  \\
      \toprule
      \multirow{2}[2]{*}{\textbf{Method}} & \multicolumn{1}{c}{\textbf{ModelNet\ to}} & \multicolumn{1}{c}{\textbf{ModelNet to}} & \multicolumn{1}{c}{\textbf{ShapeNet to}} & \multicolumn{1}{c}{\textbf{ShapeNet to}} & \multicolumn{1}{c}{\textbf{ScanNet to}} & \multicolumn{1}{c}{\textbf{ScanNet to}} & \multicolumn{1}{c}{\multirow{2}[2]{*}{\textbf{Avg}}} \\
      \multicolumn{1}{c|}{} & \multicolumn{1}{c}{\textbf{ShapeNet}} & \multicolumn{1}{c}{\textbf{ScanNet}} & \multicolumn{1}{c}{\textbf{ModelNet}} & \multicolumn{1}{c}{\textbf{ScanNet}} & \multicolumn{1}{c}{\textbf{ModelNet}} & \multicolumn{1}{c}{\textbf{ShapeNet}} &  \\
      \midrule
      No Adaptation & 80.5 & 41.6  & 75.8  & 40.0  & 60.5  & 63.6    & 60.3
      \\
      \midrule
    PointDAN \cite{Qin2019pointdan}    & 80.2  & 45.3  & 71.2  & 46.9  & 59.8  & 66.2 & 61.6 \\
    DefRec+PCM \cite{achituve2021self} & 81.1  & 50.3  & 54.3  & 52.8  & 54.0  & 69.0 & 60.3 \\
    3D Puzzle  \cite{alliegro2021joint} & 81.6  & 49.7  & 73.6  & 41.9  & 65.9  & 68.1 & 63.5 \\
    RefRec \cite{cardace2021refrec}    & 81.4  & 56.5  & \textbf{85.4}  & 53.3  & 73.0  & 73.1 & 70.5 \\
    
    (Ours)                             & \textbf{83.4}  & \textbf{61.6}  & 77.3  & \textbf{57.7} & \textbf{78.6}   & \textbf{79.8} & \textbf{73.1} \\
    
    \midrule
    Oracle                             & 93.2 & 66.2 & 95 &  66.2 & 95.0 & 93.2 \\
    \bottomrule
    \end{tabular}%
  }
\caption{Shape classification accuracy (\%) on the PointDA-10 dataset with PointNet. For each method, we report the average results on three runs. Best result on each column is in bold.}  
\label{tab:pointnet}%
\vspace{-1em}
 \end{table}%
 
\begin{table}[t]
  \centering
  \setlength{\tabcolsep}{0.8mm}
  \scalebox{0.54}{
  \begin{tabular}{p{8em}|ccccccc}
      \multicolumn{1}{c}{} &       &       &       &       &       &       &  \\
      \toprule
      \multirow{2}[2]{*}{\textbf{Method}} & \multicolumn{1}{c}{\textbf{ModelNet\ to}} & \multicolumn{1}{c}{\textbf{ModelNet to}} & \multicolumn{1}{c}{\textbf{ShapeNet to}} & \multicolumn{1}{c}{\textbf{ShapeNet to}} & \multicolumn{1}{c}{\textbf{ScanNet to}} & \multicolumn{1}{c}{\textbf{ScanNet to}} & \multicolumn{1}{c}{\multirow{2}[2]{*}{\textbf{Avg}}} \\
      \multicolumn{1}{c|}{} & \multicolumn{1}{c}{\textbf{ShapeNet}} & \multicolumn{1}{c}{\textbf{ScanNet}} & \multicolumn{1}{c}{\textbf{ModelNet}} & \multicolumn{1}{c}{\textbf{ScanNet}} & \multicolumn{1}{c}{\textbf{ModelNet}} & \multicolumn{1}{c}{\textbf{ShapeNet}} &  \\
      \midrule
      No Adaptation                      & 83.3 & 43.8 & 75.5 & 42.5 & 63.8 & 64.2 & 62.2 \\
      \midrule
      PointDAN  \cite{Qin2019pointdan}    & 83.9 & 44.8 & 63.3 & 45.7 & 43.6 & 56.4 & 56.3 \\
      DefRec+PCM  \cite{achituve2021self} & 81.7 & 51.8 & 78.6 & 54.5 & 73.7 & 71.1 & 68.6 \\
      GAST  \textsuperscript{\textdagger} \cite{Zou_2021_ICCV}     & \textbf{84.8} & 59.8 & \textbf{80.8} & 56.7 & 81.1 & 74.9 & 73.0 \\
        GLRV   \cite{Fan_2022_CVPR}                            & 85.4 & 60.4 & 78.8 & 57.7 & 77.8 & 76.2 & 72.7 \\
        ImplicitPCDA \cite{Shen_2022_CVPR}                            & 86.2 & 58.6 & 81.4 & 56.9 & 81.5 & 74.4 & 73.2 \\
      (Ours)                             & 83.9 & \textbf{61.1} & 80.3 & \textbf{58.9} & \textbf{85.5} & \textbf{80.9} & \textbf{75.1} \\
    
      \midrule
      Oracle                             & 93.9 & 78.4 & 96.2 & 78.4 & 96.2 & 93.9 & 80.5 \\
      \bottomrule
     \end{tabular}%
  }
\caption{Shape classification accuracy (\%) on the PointDA-10 dataset with DGCNN. For each method, we report the average results on three runs. Best result on each column is in bold. {\textdagger} Denotes a more powerful variant of DGCNN and results are obtained by performing checkpoint selection on the test set.}
\label{tab:dgcnn}%
 \end{table}%

\subsection{Results}
\textbf{Classification.}
We report in \cref{tab:pointnet} and \cref{tab:dgcnn} our results with PointNet and DGCNN, respectively. For PointNet, we establish overall the new state-of-the-art with 73.1\% in terms of accuracy.
We also note that our framework achieves the best results in 5 out of 6 settings, with a big gap in \mtosc{} and \stosc{} (+5.1\% and +4.4\%) that are the most challenging scenarios as they involve synthetic-to-real UDA. 
In particular, we highlight the result obtained in \mtosc{} (61.6\%), which is, roughly, only 5\% less than the oracle.
We also observe remarkable improvements when addressing the opposite case \ie{} real-to-synthetic (last two columns).
This demonstrates the capability of our framework to deal with large domain shifts. As regards as synthetic-to-synthetic UDA, we observe good performance in \mtos{}, while we are the second best model in \stom{}.
We attribute the gap with RefRec to the peculiarity of \stom{}, where the source domain is a complex dataset while the target is a simple one with shapes clearly distinguishable among classes \ie{} objects with similar shapes do belong to the same class. In such specific scenarios, reconstruction-based approaches such as RefRec shine since the auxiliary task of reconstructing a point cloud naturally tends to form well-shaped clusters in feature space that are amenable for classification.

Furthermore, we repeat the same experiments using DGCNN as our main backbone.
We achieve again state-of-the-art result (75.1\%), showing the generality of our approach towards other architectures.
Overall, we observe a similar trend w.r.t. \cref{tab:pointnet}, with an increase in performance in almost all configurations w.r.t. previous works.

\textbf{Part Segmentation.}
Although our main goal is to propose a method that aims at solving UDA for point cloud classification, our method can be easily extended to more challenging tasks such as part segmentation, which consists in assigning to each vertex of the shape one object category. 
As done for point cloud classification, we perform a first step of \sd{} to distill good features for the target domain unsupervisedly. We then simply adapt the \st{} step by considering each vertex of the input shape as a node in the graph. In this case, the node representation consists of a local feature vector extracted from the main backbone, which is a PointNet as in \cite{alliegro2021joint}. The whole graph is theoretically composed of all points of all shapes in the dataset. However, keeping all vertices in memory would be impractical and we perform the procedure explained in \cref{subsec:fd} by considering 20000 points of the whole dataset for each refinement iteration.
Results are reported in \cref{tab:partseg}. The evaluation metric is the mean Intersection-over-Union (mIoU), which is computed for each part Q for all the samples of the \textit{chair} class. Then, the average across parts is reported.
First, we observe that our full framework (last row) surpasses by more than 10\% the previous method (second row). Furthermore, we highlight the effectiveness of \sd{} for the part segmentation task. Indeed, when only performing the first step of our pipeline (third row of \cref{tab:partseg}), we already overcome \cite{alliegro2021joint} by 
7.7\%.

\begin{table}
    \centering
    \setlength{\tabcolsep}{5.8mm}
    \scalebox{0.65}{
   \begin{tabular}{l|cccc|c@{}}
    \hline
    \multicolumn{6}{c}{Part Segmentation: ShapeNetPart $\rightarrow$ ScanOBJ\_BG} \\
        \hline
            \footnotesize{Method} & \footnotesize{Seat} & \footnotesize{Back} & \footnotesize{Base} & \footnotesize{Arm} & \footnotesize{Avg.} \\
        \hline
            Source only & 67.85 & 45.60 & 84.89 & 14.87 & 53.30  \\
            3D Puzzle \cite{alliegro2021joint} & 65.70 & 49.11 & 85.91 & 21.40 & 55.53  \\
        \hline        
            Self-dist (ours) & 71.1 & 79.3 & 65.2 & 37.0 & 63.2  \\
            (ours) & \textbf{74.7} & \textbf{82.7} & \textbf{67.9} & \textbf{37.7} & \textbf{65.7}  \\
        \hline
    \end{tabular}
    }
\caption{Per part and average mIoU (\%) of chair segmentation for ShapeNetPart to ScanOBJ-BG.}  
\label{tab:partseg}%
\end{table}

\begin{table}[t]
   \centering
  \setlength{\tabcolsep}{2mm}
  \scalebox{0.46}{
   \begin{tabular}{p{3em}|cccccccccc}
      \multicolumn{1}{c}{} & & & &       &       &       &       &       &       &  \\
      \toprule
      \multirow{2}[2]{*}{\textbf{Step}} & \multirow{2}[2]{*}{ce} & \multirow{2}[2]{*}{sd} & \multirow{2}[2]{*}{kd} & \multicolumn{1}{c}{\textbf{ModelNet\ to}} & \multicolumn{1}{c}{\textbf{ModelNet to}} & \multicolumn{1}{c}{\textbf{ShapeNet to}} & \multicolumn{1}{c}{\textbf{ShapeNet to}} & \multicolumn{1}{c}{\textbf{ScanNet to}} & \multicolumn{1}{c}{\textbf{ScanNet to}} & \multicolumn{1}{c}{\multirow{2}[2]{*}{\textbf{Avg}}} \\
      \multicolumn{1}{c|}{} & & & & \multicolumn{1}{c}{\textbf{ShapeNet}} & \multicolumn{1}{c}{\textbf{ScanNet}} & \multicolumn{1}{c}{\textbf{ModelNet}} & \multicolumn{1}{c}{\textbf{ScanNet}} & \multicolumn{1}{c}{\textbf{ModelNet}} & \multicolumn{1}{c}{\textbf{ShapeNet}} &  \\
      \midrule
      \multirow{3}{*}{PL init} & \multicolumn{1}{c|}\checkmark & \multicolumn{1}{c|}{} & \multicolumn{1}{c|}{} & 80.5 & 41.6 & 75.8 & 40.0  & 60.5  & 63.6    & 60.3
      \\
        & \multicolumn{1}{c|}\checkmark & \multicolumn{1}{c|}\checkmark & \multicolumn{1}{c|}{} &  \textbf{82.1} & \textbf{57.2} & 77.6 & \textbf{55.0} & \textbf{71.0} & \textbf{72.1}  & \textbf{69.2} \\
       & \multicolumn{1}{c|}\checkmark & \multicolumn{1}{c|}{} & \multicolumn{1}{c|}\checkmark & 79.6 & 54.0 &  \textbf{79.2} & 53.2 & 53.9 & 70.0  & 65.0 \\
      \bottomrule
     \end{tabular}%
   }
\caption{Ablation study for the first step of our framework. ce: cross-entropy loss on source domain sd: self-distillation loss \cref{eq:dino_loss} in feature space used to train the \pl{} model; kd: standard knowledge distillation loss \cite{hinton2015distilling} in output space. We report the average results on three runs.}  
\label{tab:distillation}%
\vspace{-1em}
 \end{table}%
 
\textbf{Self-distillation vs knowledge distillation.}
In \cref{tab:distillation}, we ablate our self-distillation strategy and also compare it to an obvious alternative, \ie{} applying \cref{eq:dino_loss} in output space.
In this case, the \sd{} loss in \cref{eq:dino_loss} is applied on the output of the classifier rather than the feature vector of the backbone.
As explained in \cref{sec:related}, this protocol is similar to the knowledge distillation paradigm \cite{hinton2015distilling} that uses soft \pl{}.
While we observe in both cases an improvement over the baseline trained only on source data (first row), the improvement is twice as large when \sd{} is deployed, which demonstrates the importance of working in feature space. Moreover, the large improvement in absolute terms (+8.9\% on average) attained by using \sd{} shows its effectiveness in reducing the domain gap, validating our intuition to use it to tackle UDA. 
Interestingly, we observe a different behaviour for \stom{}. This is again likely due to the peculiarity of the setting. With the source domain being much larger and richer than the target one, it is plausible that \pl{} in output space are quite accurate, end therefore more effective in this case.
The model trained with \sd{} is used to extract the initial set of \pl{} for our method, as well as for all the \st{} variants compared in \cref{tab:ablation}.
Finally, we highlight how the results obtained with \sd{} are clearly superior in all scenarios on average to those attained by competitors based on self-supervised learning tasks, \eg{} row 2 (DefRec) and 3 (3D puzzle) of \cref{tab:pointnet}, that are based on a reconstruction and a 3D puzzle pretext task, respectively.
This provides empirical support for our claim on the higher effectiveness of \sd{} with respect to auxiliary tasks for 3D UDA.

\begin{table}[t]
   \centering
  \setlength{\tabcolsep}{2mm}
  \scalebox{0.45}{
   \begin{tabular}{p{5em}|ccccccccccc}
      \multicolumn{1}{c}{} & & & & & & & & & &   \\
      \toprule
      \multirow{2}[2]{*}{\textbf{Step}}  & \multirow{2}[2]{*}{st} & \multirow{2}[2]{*}{ref} &\multirow{2}[2]{*}{sd} & \multicolumn{1}{c}{\textbf{ModelNet\ to}} & \multicolumn{1}{c}{\textbf{ModelNet to}} & \multicolumn{1}{c}{\textbf{ShapeNet to}} & \multicolumn{1}{c}{\textbf{ShapeNet to}} & \multicolumn{1}{c}{\textbf{ScanNet to}} & \multicolumn{1}{c}{\textbf{ScanNet to}} & \multicolumn{1}{c}{\multirow{2}[2]{*}{\textbf{Avg}}} \\
      \multicolumn{1}{c|}{} & \multicolumn{3}{c}{} & \multicolumn{1}{c}{\textbf{ShapeNet}} & \multicolumn{1}{c}{\textbf{ScanNet}} & \multicolumn{1}{c}{\textbf{ModelNet}} & \multicolumn{1}{c}{\textbf{ScanNet}} & \multicolumn{1}{c}{\textbf{ModelNet}} & \multicolumn{1}{c}{\textbf{ShapeNet}} &  \\
      \midrule
      \multirow{3}{*}{Adaptation}       & \multicolumn{1}{c|}\checkmark & \multicolumn{1}{c|}{}& \multicolumn{1}{c|}{} & 82.7 & 59.3 & 74.9 & 56.4 & 77.1 & 77.8 & 71.4 \\
           & \multicolumn{1}{c|}\checkmark & \multicolumn{1}{c|}{\checkmark} & \multicolumn{1}{c|}{}         & \textbf{83.4} & 60.9 & \textbf{78.2} & 56.3 & 77.9 & 79.4 & 72.7   \\
           & \multicolumn{1}{c|}\checkmark & \multicolumn{1}{c|}{\checkmark} & \multicolumn{1}{c|}{\checkmark}         & \textbf{83.4} & \textbf{61.6} & 77.3 & \textbf{57.7} & \textbf{78.6} & \textbf{79.8} & \textbf{73.1} \\
      \bottomrule
     \end{tabular}%
   }
\caption{Ablation for the second step of our algorithm. st: \st{} with \pl{} of the last row of \cref{tab:distillation} model; sd: self-distillation loss in the adaptation step; ref: refinement of \pl{} with GCN. We average results on three runs.}  
\label{tab:ablation}%
\vspace{-1em}
 \end{table}%

\textbf{Self-training strategies.}
In \cref{tab:ablation}, we perform an ablation study on the second step of our pipeline.
We start by applying the simplest strategy to perform \st{} (first row), \ie{} using all \pl{} of the target domain together with the labels from the source domain to train a single classifier. This provides competitive results (71.4\%), that is already better than the previous state-of-the-art model (70.5\%), again showcasing the effectiveness of \sd{} to obtain \pl{} for UDA.
When activating also the proposed online refinement that iteratively improves \pl{} thanks to the global reasoning of the GCN (second row), we appreciate another large improvement compared to the naive \st{}, which validates the importance of the proposed iterative refinement. 
Finally, in the last row, we report the results attained by activating \sd{} also in the adaptation step, which leads to the best performance and is the model used in all other experiments.
\begin{figure}[t]%
    \centering
    {\includegraphics[scale=0.45]{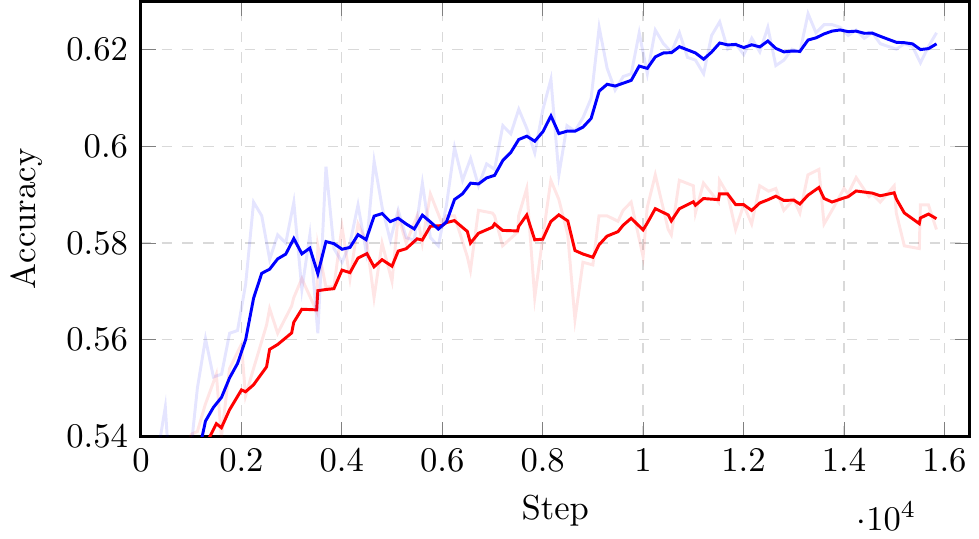} }%
    \caption{Test accuracy on target domain during training on \mtosc{}.
	Our model (Blue) consistently improves \pl{} during training differently from a simple \st{} strategy in which \pl{} are fixed (Red).}
	\vspace{-1em}
    \label{fig:training}
\end{figure}
As a further validation of the importance of the design decisions in our framework, we plot the training curves of the synthetic-to-real \mtosc{} in \cref{fig:training}. The curves represent the test accuracy on the target domain during training. The red plot shows the behaviour of the naive \st{}, which correspond to row 1 of \cref{tab:ablation}. On the other hand, the blue lines represent the training curves obtained with our full model, \ie{} last row of \cref{tab:ablation}. We can appreciate that, after a certain number of steps, the blue line is always above the red line. This is a clear evidence that in our full model, \pl{} are improved over time, while in the naive case the model starts to overfit, leading to a plateau. We also wish to point out that such behaviour is key to a good UDA method because, in absence of target labels to perform validation, it is basically impossible to decide when to stop the training process.

%% file: sections/04-conclusions.tex
\section{Limitations}
\label{sec:limitation}
The main limitation of the proposed approach is the hand-crafted data augmentation functions used to augment both source and target data. To this end, we would like to investigate the possibility to learn a transformation able to automatically model the gap between the two domains. This would allow to handle dynamically cases where less augmentation is needed, such as in \stom{}.

\section{Conclusion}
\label{sec:conclusion}
In this work, we explored a novel strategy to learn features on the target domain without the need of annotations. We first proposed to guide the network to learn a clustered feature space for the target domain and preserve discriminability suitable for classification. In addition, we introduced a novel refinement strategy that is able to globally reason on the target domain by means of GNN and to correct misclassified samples during training.
Combining the two contributions, allowed to establish the state-of-the-art in the reference benchmarks.
Finally, we showed how these contributions can be used for more challenging tasks such as part segmentation.